\documentclass[conference]{IEEEtran}

\usepackage{amsmath,amssymb}
\usepackage{booktabs}
\usepackage{graphicx}
\usepackage{multirow}
\usepackage{cite}
\usepackage{url}
\usepackage{xcolor}
\usepackage{microtype}
\usepackage{newtxtext,newtxmath}
\usepackage{balance}

\definecolor{RouteBlue}{RGB}{35,86,145}
\newcommand{\method}{\textsc{HRD}}

\title{Beyond Final-Token Classification: Heterogeneous Readouts for Evidence-Grounded Suicide Risk Detection}

\author{
\IEEEauthorblockN{Zirui Li and Yanling Li}
\IEEEauthorblockA{
The Hong Kong Polytechnic University, Hong Kong SAR, China\\
25044237g@connect.polyu.hk; lynnn.li@connect.polyu.hk}
\and
\IEEEauthorblockN{Kaolanglang Gao}
\IEEEauthorblockA{
Shenzhen University, Shenzhen, China\\
2510032049@mails.szu.edu.cn}
}

\begin{document}
\maketitle

\begin{abstract}
The IEEE BigData Cup benchmark combines three prediction problems with different output structures: ordinal suicide-risk classification, multi-label psychosocial factor detection, and extraction of supporting phrases. We introduce heterogeneous readout decomposition (\method{}), which separates semantic verification from output realization. A locally deployed Qwen3.8-27B model, adapted with task-specific QLoRA adapters, produces both answer-token margins and layer-63 answer states for card-conditioned queries. \method{} compares four latent scores for ordinal risk, retains token margins for most factors while routing seven labels through one shared latent probe, and constructs evidence sets from verbatim span candidates with calibrated, risk-conditional constraints. On two held-out user-grouped confirmation folds, the latent risk readout improves weighted F1 from 0.8237 to 0.8372 and macro F1 from 0.7965 to 0.8185. Selective factor routing improves macro F1 by 0.0105 and tail-label macro F1 by 0.0189; in contrast, global latent replacement and independent label-specific probes fail. The constrained evidence decoder raises pooled phrase F1 from 0.7488 to 0.7609 in row-level out-of-fold evaluation. Lenormand's best public result is 0.8052 on Subtask~1 and 0.6636 on Subtask~2, giving a composite score of 0.7627. These results identify the answer-token boundary, rather than semantic representation alone, as a measurable source of error in this benchmark.
\end{abstract}


\section{Introduction}
Social media posts can contain early signals of suicidal ideation, planning, attempts, and the psychosocial conditions surrounding them. Computational systems that analyze these posts must therefore do more than assign a risk label: their outputs should be grounded in the author's words and should distinguish risk factors from protective resources. The Explainable Suicide Risk Detection challenge at IEEE BigData Cup 2026 operationalizes this requirement through three outputs for each Reddit post: one of four C-SSRS-grounded risk levels, one or more verbatim evidence phrases, and a set of 24 risk or protective factors. The risk scale and factor ontology draw on established clinical and psychosocial constructs~\cite{li2025protective,posner2011cssrs}.

Although all three outputs are predicted from the same post, they have different statistical structure. Risk is an ordinal, mutually exclusive decision. Evidence is a variable-size set of substrings, and small boundary changes affect the official score. Factors form a long-tailed multi-label ontology whose neighboring constructs, including \emph{poor social support} and \emph{interpersonal difficulty}, often share lexical cues. An unconstrained generation prompt does not encode these distinctions. Autoregressive generation can suppress additional labels that remain plausible to the model~\cite{ma2025multilabel}, and generated explanations need not be verbatim evidence~\cite{chim2024overview,tran2024ism}.

Our starting point is a semantic verifier rather than a conventional 24-way output head. Each risk level and factor is represented by an operational card with a definition, inclusion and exclusion rules, and common confusions. The 27B model judges each card independently. This gives the labels a common semantic interface, but the final \texttt{YES/NO} or \texttt{PRESENT/ABSENT} token is still only one possible readout of the verifier state. Tuned-lens studies show that transformer states admit informative auxiliary predictions~\cite{belrose2023tunedlens}; RISE uses internal classifiers for robust suicide-risk early exit~\cite{soun2024rise}. We instead test whether the final hidden state supports a better task boundary than the answer-token margin, and whether that advantage is consistent across labels.

The resulting heterogeneous readout decomposition (\method{}) uses a shared 4-bit Qwen3.8-27B backbone with separate task adapters. Its central object is a \emph{verification record}: the answer-token margin and answer-position hidden state produced for a post--card pair. Output-specific compilers then transform these records into a four-way ordinal comparison, 24 independent factor decisions, or a constrained set of exact spans. Thus the contribution is not a larger classifier or an additional generation stage; it is a controlled separation between what the verifier represents and how each output space reads that representation. Risk is read from four card-conditioned layer-63 states. Factors retain answer-token margins for 17 labels and use one shared latent probe for seven development-selected labels. Evidence does not use free-form generation: a proposer enumerates exact spans, the verifier scores them, and a logistic calibrator constructs a risk-conditioned evidence set.

This study makes three contributions:
\begin{itemize}
    \item \textbf{A task-structural formulation.} We cast risk, factor, and evidence prediction as a common card-conditioned verification step followed by three output-specific compilers. This formulation exposes the final token as one readout choice rather than treating it as the semantic decision itself.
    \item \textbf{A selective latent-routing algorithm.} We combine answer-token margins and a regularized shared probe through a frozen label route. The comparison is falsifiable: global latent replacement and seven independent probes fail, whereas the shared seven-label route improves factor macro F1 on both confirmation folds and the four-card latent comparator improves risk weighted F1.
    \item \textbf{Metric-aligned grounding and validation.} We decode only exact input spans under one-to-one event constraints, and we separate development, frozen confirmation, row-level OOF evidence evaluation, and public-leaderboard results. This couples the method to the benchmark's output geometry while making the strength of each empirical claim explicit.
\end{itemize}

\section{Task, Data, and Evaluation}
\subsection{Task Definition}
Subtask~1 predicts \emph{Indicator}, \emph{Ideation}, \emph{Behavior}, or \emph{Attempt}. The official risk metric is weighted F1. Evidence is a semicolon-separated set of verbatim phrases and is evaluated by phrase F1 after case-insensitive normalization. A predicted phrase matches a gold phrase when either contains the other, subject to a three-times token-length cap and one-to-one matching. Subtask~2 predicts any subset of 24 factors and is evaluated by macro F1. The overall competition score is
\begin{equation}
S=0.4F_{\mathrm{risk}}+0.3F_{\mathrm{evidence}}+0.3F_{\mathrm{factor}}.
\end{equation}
The leaderboard reports a normalized Subtask~1 score and the factor score separately.

\subsection{Dataset}
We use only the official training set: 1,635 posts from 153 anonymized users. It contains 611 Indicator, 519 Ideation, 391 Behavior, and 114 Attempt instances. The median post length is 38 whitespace-delimited tokens, the mean is 68.8, and the maximum is 2,066. After de-duplicating repeated factor annotations within a post, each post has 2.92 factors on average. Label support ranges from 745 instances for \emph{hopelessness} to eight for \emph{sexual orientation related issues}. This imbalance makes accuracy and micro F1 inadequate for factor model selection.

We form three user-grouped folds of sizes 548, 543, and 544. All posts from a user remain in one fold. Retrieval indices, semantic caches that use training labels, thresholds, and probe fitting respect this partition. No external labeled examples are used. Publicly available pretrained models and manually authored operational label cards are declared components of the system.

\section{Related Work}
\subsection{Suicide Risk and Evidence Grounding}
Prior work has used prompting, pseudo-labeling, and model ensembles when suicide labels are limited~\cite{nguyen2024limited}. The CLPsych 2024 shared task specifically studied evidence of suicide risk in online posts~\cite{chim2024overview}; participating systems included knowledge self-generation and output refinement with an LLM~\cite{tran2024ism}. In contrast, our evidence decoder is non-generative: every returned phrase is an exact substring of the input. The 24-factor taxonomy is adapted from work that jointly models risk and protective factors over time~\cite{li2025protective}. We use these constructs as annotation variables, not as clinical diagnoses.

\subsection{Label Semantics and Multi-label Prediction}
Natural-language label descriptions can transform classification into entailment between an input and a label hypothesis~\cite{talavera2023labelaware}. This is particularly relevant for long-tailed ontologies because parameters can be shared across labels while their semantics remain explicit. Recent analysis further shows that autoregressive multi-label generation behaves differently from independent classification and may suppress otherwise plausible labels~\cite{ma2025multilabel}. We consequently ask one binary question per factor and read a continuous margin rather than generate a factor list.

\subsection{Internal Readouts and Selective Routing}
The tuned lens decodes predictions from frozen transformer blocks through learned affine maps~\cite{belrose2023tunedlens}. RISE trains robust internal classifiers for suicide risk and uses them for ensembling, abstention, and early exit~\cite{soun2024rise}. Our work is complementary: the deployed model still runs to its final layer, and we study a mismatch between the final hidden state and the final answer-token boundary. Label-selective routing is conceptually related to learning to defer~\cite{mozannar2020defer}, but our ``experts'' are two readout interfaces of one model and the route is frozen at the label level rather than selected per test instance. Finally, probe accuracy alone does not establish that a feature is causally used by the model~\cite{hewitt2019probes,belinkov2022probing}; we therefore use the term \emph{predictively accessible} and do not interpret probe weights as psychological mechanisms.

\subsection{Extractive Rationales}
Extractive rationale models separate selection from prediction~\cite{lei2016rationalizing} and can enforce faithfulness by construction when the predictor is restricted to selected text~\cite{jain2020fresh}. Sufficiency, comprehensiveness, and compactness regularization have also been studied for multi-label rationale extraction~\cite{chalkidis2021rationale}. Our exact-copy constraint prevents fabricated evidence, but the risk model can still access the full post. We therefore claim verbatim grounding, not causal faithfulness.

\section{Method}
\begin{figure*}[t]
    \centering
    \includegraphics[width=0.98\textwidth]{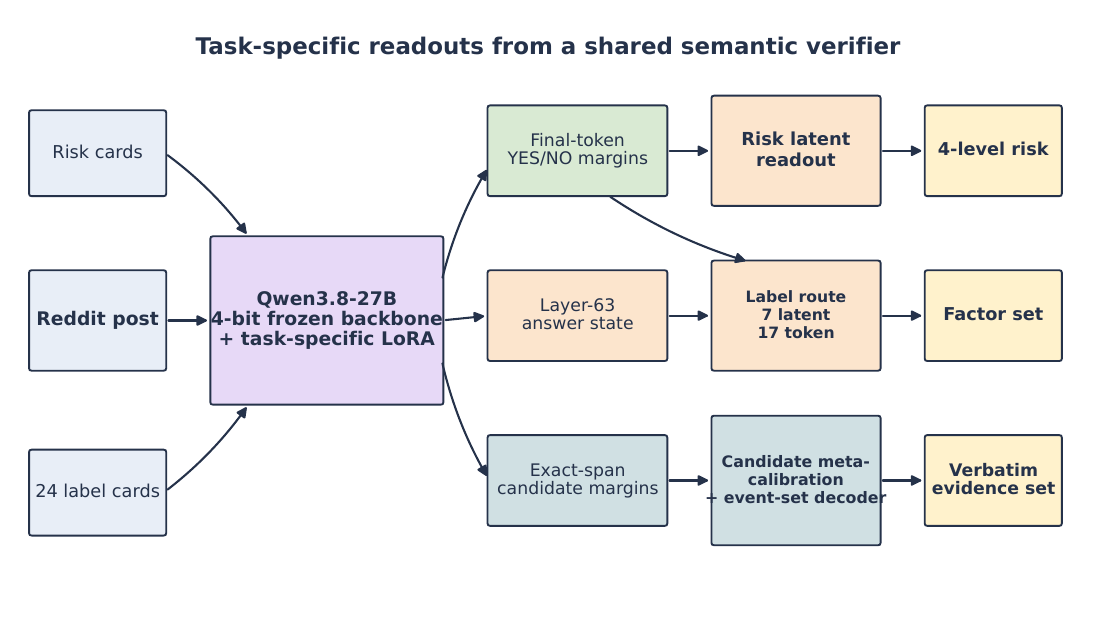}
    \caption{Overview of \method{}. Operational cards create a common semantic-verification interface, while risk, factor, and evidence outputs use different readouts. The factor route is label selective and frozen before confirmation.}
    \label{fig:architecture}
\end{figure*}

Figure~\ref{fig:architecture} summarizes the complete system. Qwen3.8-27B is a 64-layer hybrid linear/full-attention model~\cite{qwen38modelcard}. We load the backbone in 4-bit NF4 with bfloat16 computation and train task-specific low-rank adapters following QLoRA~\cite{dettmers2023qlora}. Risk/evidence and factors use separate adapters because their supervision and prompts differ; parameters are shared across all labels within each task.

For post $x_i$ and operational card $c_l$, a task adapter produces a verification record
\begin{equation}
\mathcal{V}_{\theta}(x_i,c_l)=\left(m_{i,l},h_{i,l}^{(63)}\right),
\end{equation}
containing the positive--negative answer-token margin and the final answer-position state. \method{} applies an output compiler $\mathcal{R}_t$ appropriate to task structure:
\begin{equation}
\hat{y}^{(t)}_i=\mathcal{R}_t\!\left(\{\mathcal{V}_{\theta}(x_i,c_l)\}_{l},C(x_i)\right),
\end{equation}
where $C(x_i)$ is used only by the extractive evidence compiler. The following subsections instantiate $\mathcal{R}_t$ for ordinal risk, independent factors, and evidence sets.

Table~\ref{tab:hrdalgorithm} gives the complete inference rule. The route $\rho_l$, probe weights, candidate calibrator, and thresholds are fitted before confirmation and remain fixed at test time.

\begin{table}[t]
\caption{Heterogeneous readout decomposition at inference.}
\label{tab:hrdalgorithm}
\centering
\small
\begin{tabular}{@{}lp{0.82\columnwidth}@{}}
\toprule
Step & Operation \\
\midrule
1 & For every relevant card $c_l$, compute $\mathcal V_\theta(x,c_l)=(m_l,h_l^{(63)})$. \\
2 & Risk: jointly compare four probe scores $q_r$ and return $\arg\max_r q_r$. \\
3 & Factor $l$: use $g_l$ when the frozen route $\rho_l=1$; otherwise retain margin $m_l$. Apply its OOF threshold independently. \\
4 & Evidence: score only $c\in C(x)$, calibrate candidate probabilities, then deduplicate and apply the risk-conditional set cap. \\
\midrule
Output & One risk level, a factor subset, and exact-copy evidence phrases. \\
\bottomrule
\end{tabular}
\end{table}

\subsection{Operational Cards as a Semantic Interface}
For a post $x_i$ and label $l$, a card $c_l$ contains a name, operational definition, inclusion rule, exclusion rule, and confusable labels. A factor prompt has the form
\begin{quote}\small
Does the post support the supplied factor?\\
\texttt{A = PRESENT; B = ABSENT.}
\end{quote}
Risk uses four analogous one-vs-rest cards with \texttt{YES/NO} answers. The final-token margin is
\begin{equation}
m_{i,l}=z_{i,l}^{+}-z_{i,l}^{-},
\end{equation}
where $z^{+}$ and $z^{-}$ are the logits for the positive and negative answer tokens at the answer position.

For long posts, we preserve the opening and ending clauses and add the clauses most similar to the card embedding. E5-large-v2 embeddings~\cite{wang2022e5} provide this relevance score. Factor prompts additionally include one fold-safe positive precedent and one semantically similar negative precedent. Eligibility excludes the validation fold, the current user, exact duplicate text, and the query itself. Negatives prioritize boundary confusions specified on the card. This retrieval is used as context during SFT and scoring; it is not a post-hoc label vote.

\subsection{Risk Latent Readout}
Let $h_{i,r}^{(k)}$ be the answer-position state from decoder layer $k$ when post $i$ is queried with risk card $r$. We treat the four cards as binary pairs and fit a shared logistic probe
\begin{equation}
q_{i,r}=w^\top [h_{i,r}^{(k)}; e_r]+b,
\end{equation}
where $e_r$ is a four-dimensional risk identity vector. The predicted risk is $\arg\max_r q_{i,r}$. Standardization and logistic regression are fit only on the outer-training users.

Layers $\{15,31,47,63\}$ and $C\in\{0.001,0.01,0.1\}$ were evaluated through inner grouped cross-validation on development Fold~0. Layer 63 (zero indexed) and $C=0.01$ were frozen, then evaluated without retuning on Folds~1 and 2. The confirmation experiment in Sec.~VI-B evaluates the full latent decision. For the public submission, we used a more conservative test-time rule: an Indicator prediction was changed to Ideation only when all three fold-specific probes predicted Ideation and their mean Ideation-minus-Indicator probability margin was at least 0.08. This frozen rule changed one test row and left the evidence and factor fields unchanged.

\subsection{Label-Selective Factor Readout}
A global latent factor probe uses
\begin{equation}
g_{i,l}=v^\top[h_{i,l}^{(63)};m_{i,l};e_l]+a,
\end{equation}
where $e_l$ is a 24-dimensional label identity vector. The probe is deliberately shared rather than fitting 24 independent heads. It adds only a standardized logistic readout to the frozen 27B representation.

The latent interface is not uniformly better. On development Fold~0, global replacement raises average precision but changes macro F1 by only $+0.0002$ and reduces tail macro F1. We therefore freeze a label route $\rho_l$:
\begin{equation}
s_{i,l}=\begin{cases}
g_{i,l}, & \rho_l=1,\\
m_{i,l}, & \rho_l=0.
\end{cases}
\end{equation}
Seven labels pass the development gate: \emph{psychological capital}, \emph{social support}, \emph{poor school performance}, \emph{sense of responsibility}, \emph{traumatic experience}, \emph{low socio-economic status}, and \emph{interpersonal difficulty}. Layer 63, $C=0.001$, the route set, and all threshold procedures are then fixed. Folds~1 and 2 are used only for confirmation. This is a static label route, not a sample-adaptive expert selector.

The base factor score is a probability blend of a Qwen3-14B shared verifier~\cite{qwen3report} (weight 0.25) and the Qwen3.8-27B verifier (weight 0.75). The 14B component is retained because its errors are not perfectly correlated with the 27B model. Per-label thresholds are estimated from user-grouped out-of-fold predictions. The latent route replaces only the seven designated label scores before thresholding.

\subsection{Constrained Evidence-Set Decoding}
Evidence extraction is formulated as selection from a candidate inventory $C(x)$ rather than token generation. The inventory combines sentence and clause boundaries, cue-centered word windows, a fold-trained suicide lexicon, and token regions proposed by a ModernBERT encoder~\cite{warner2025modernbert}. Every candidate records its exact character offsets and therefore remains a verbatim substring.

The Task~1 adapter receives the provisional risk card, the post, and one candidate span, and returns a support margin. A low-capacity logistic calibrator estimates candidate relevance from the frozen margin and test-time-observable metadata: candidate source, three predicted risk fields, span length, relative position, candidate count, margin rank, distance from the top margin, baseline phrase count, and post length. It does not access gold evidence at inference time.

Candidate probabilities are converted into an evidence set by a frozen threshold of 0.40, exact-string deduplication, and suppression of overlapping events within 40 characters. Risk-conditional caps return zero phrases for Indicator, at most two for Ideation, three for Behavior, and two for Attempt. This prevents the common failure mode of maximizing recall by returning long or redundant spans and mirrors the official one-to-one phrase matching rule.

\section{Experimental Protocol}
\subsection{Development and Confirmation}
\begin{figure}[t]
    \centering
    \includegraphics[width=\columnwidth]{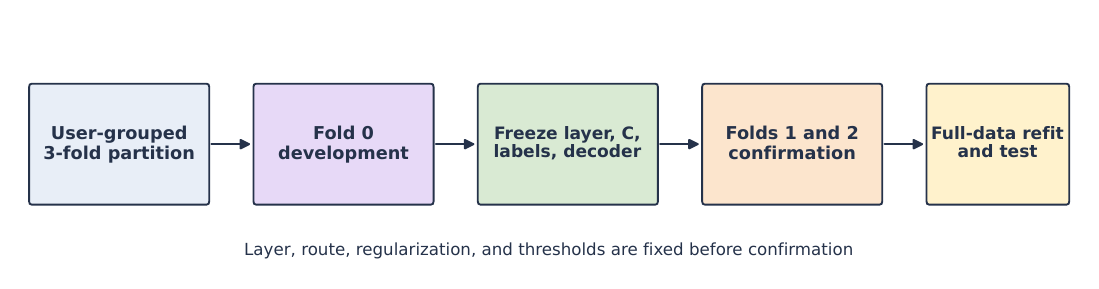}
    \caption{Selection protocol for latent readouts. Confirmation folds do not choose the layer, regularization, route, or threshold.}
    \label{fig:protocol}
\end{figure}

Our large design space makes ordinary pooled cross-validation susceptible to selection leakage. We therefore separate \emph{development} from \emph{confirmation}, as shown in Fig.~\ref{fig:protocol}. Fold~0 selects the hidden layer, regularization, route labels, and decoder hyperparameters. These choices are serialized before Folds~1 and 2 are evaluated. A component is accepted only if it satisfies a pre-declared pooled gain and does not rely on one favorable fold.

Risk acceptance requires pooled weighted-F1 gain of at least 0.006, nonnegative pooled macro-F1 gain, no fold below $-0.003$ weighted F1, and paired-bootstrap probability of positive gain above 0.90. Factor routing requires at least $+0.006$ pooled macro F1 and macro AP, tail macro F1 above a $-0.005$ floor, and strictly positive macro-F1 gain on each confirmation fold.

The evidence experiment has a weaker methodological status. Its candidate-level calibrator is frozen before outer Fold~1/2 evaluation, and predictions are row-level out of fold. However, the Fold~0 candidate generator was trained using Folds~1/2. Thus evidence numbers are valid for competition model selection but are not a fully nested external-test estimate. We preserve this distinction rather than describing all three components as equivalently untouched.

\subsection{Implementation}
Both 27B adapters use one epoch, learning rate $10^{-4}$, rank 16, LoRA scale 32, dropout 0.05, maximum length 1,536, batch size 1, and gradient accumulation 32. All 64 language layers are adapted. Training uses 4-bit NF4 with double quantization, bfloat16 computation, gradient checkpointing, FlashAttention-2 for full-attention blocks, and optimized causal-convolution/linear-attention kernels. Experiments run locally in Google Colab Pro on a single NVIDIA A100 80GB. The factor verifier uses balanced positive/negative pairs with a 0.7 hard-negative fraction. Random seed 42 is fixed for model and probe training.

\section{Results}
\subsection{Public Leaderboard}
Table~\ref{tab:leaderboard} gives the score of the submitted system. The leaderboard does not reveal gold labels or separate risk/evidence components, so it is used only as an external distribution check. All method selection reported below uses training-set folds.

\begin{table}[t]
\caption{Best recorded public-leaderboard result.}
\label{tab:leaderboard}
\centering
\begin{tabular}{@{}lcccc@{}}
\toprule
System & Task 1 & Task 2 & Composite & Rank \\
\midrule
Lenormand (\method{}) & \textbf{0.8052} & \textbf{0.6636} & \textbf{0.7627} & 9 \\
\bottomrule
\end{tabular}
\end{table}

\subsection{Risk Readout}
Table~\ref{tab:risk} reports the frozen confirmation. The latent readout improves weighted F1 on both folds, yielding $+0.0135$ pooled weighted F1 and $+0.0219$ macro F1. A paired post-level bootstrap gives a 95\% interval of $[0.0012,0.0261]$ for weighted-F1 gain and probability 0.983 that the gain is positive. Behavior recall rises from 0.7315 to 0.7704; Attempt recall rises from 0.6712 to 0.7397. Mean absolute severity error falls from 0.2052 to 0.1858.

\begin{table}[t]
\caption{Risk results on frozen confirmation folds.}
\label{tab:risk}
\centering
\begin{tabular}{@{}lccc@{}}
\toprule
System & Fold 1 WF1 & Fold 2 WF1 & Pooled MF1 \\
\midrule
Final-token baseline & 0.8214 & 0.8260 & 0.7965 \\
Layer-63 readout & \textbf{0.8387} & \textbf{0.8355} & \textbf{0.8185} \\
\midrule
Difference & +0.0172 & +0.0095 & +0.0219 \\
\bottomrule
\end{tabular}
\end{table}

\subsection{Factor Readout}
Table~\ref{tab:factor} compares the frozen 0.25/0.75 token blend with the seven-label route on Folds~1/2. Routing improves macro AP by 0.0082, macro F1 by 0.0105, and tail macro F1 by 0.0189. The gain is positive on both folds. Head-label macro F1 is unchanged because none of the pooled head labels is routed.

\begin{table}[t]
\caption{Pooled factor results on Folds 1 and 2.}
\label{tab:factor}
\centering
\begin{tabular}{lcccc}
\toprule
System & MAP & MF1 & Tail MF1 & Micro F1 \\
\midrule
Token blend & 0.7376 & 0.6712 & 0.5801 & 0.7553 \\
Selective route & \textbf{0.7458} & \textbf{0.6816} & \textbf{0.5990} & \textbf{0.7587} \\
\midrule
Difference & +0.0082 & +0.0105 & +0.0189 & +0.0034 \\
\bottomrule
\end{tabular}
\end{table}

Fold~1 macro F1 rises from 0.6619 to 0.6736; Fold~2 rises from 0.6788 to 0.6895. The readout advantage is therefore label dependent: selected factors benefit from the answer state, while the token boundary remains preferable for the other labels.

\subsection{Where Does Factor Routing Help?}
Table~\ref{tab:routediagnostic} decomposes the frozen route by label. Three labels improve on both confirmation folds: \emph{poor school performance}, \emph{psychological capital}, and \emph{traumatic experience}. Their supports span tail and mid-frequency strata, so frequency alone does not explain the gain. The remaining routed labels have mixed fold-level effects even though the aggregate route improves both folds. This rules out a broad claim that latent states are uniformly better for rare or abstract constructs. The supported interpretation is narrower: a shared latent correction improves particular decision boundaries, while development-only route selection remains noisy at the individual-label level.

\begin{table}[t]
\caption{Per-label F1 change from the frozen factor route. Labels were selected on Fold 0; no confirmation label was removed.}
\label{tab:routediagnostic}
\centering
\small
\begin{tabular}{@{}lrr@{}}
\toprule
Routed factor & Fold 1 & Fold 2 \\
\midrule
Poor school performance & +0.1556 & +0.1616 \\
Psychological capital & +0.0244 & +0.0249 \\
Traumatic experience & +0.0503 & +0.0655 \\
Interpersonal difficulty & -0.0085 & +0.0026 \\
Low socioeconomic status & +0.0315 & -0.0455 \\
Sense of responsibility & +0.0571 & -0.0151 \\
Social support & -0.0309 & +0.0610 \\
\bottomrule
\end{tabular}
\end{table}

\subsection{Evidence Decoder}
The frozen candidate calibrator improves pooled phrase F1 from 0.7488 to 0.7609 (Table~\ref{tab:evidence}). Its main effect is to trade redundant recall for precision: mean predicted phrases fall from 1.35 to 1.01, while phrase precision rises from 0.7237 to 0.7723. Fold~1 improves by 0.0234 and Fold~2 by 0.0007. The positive but unequal fold gains motivate the methodological qualification in Section~V-A.

\begin{table}[t]
\caption{Row-level OOF evidence results on Folds 1 and 2.}
\label{tab:evidence}
\centering
\begin{tabular}{lcccc}
\toprule
Decoder & Precision & Recall & Phrase F1 & Phrases \\
\midrule
Indicator-empty baseline & 0.7237 & \textbf{0.8178} & 0.7488 & 1.35 \\
Meta event set & \textbf{0.7723} & 0.7804 & \textbf{0.7609} & 1.01 \\
\bottomrule
\end{tabular}
\end{table}

\subsection{Falsification and Ablation}
Table~\ref{tab:falsification} records the main rejected alternatives. An additional hard-error SFT epoch increases factor ranking AP but sharply reduces thresholded macro F1. Global factor latent replacement also improves AP without improving macro F1. Pairwise ranking and structured CoT do not repair the boundary, and a risk event compiler lowers the projected Task~1 score. The seven-label shared probe is the factor readout that passes frozen confirmation.

\begin{table*}[t]
\caption{Falsification-oriented ablations. ``Dev'' denotes Fold~0 screening; ``OOF'' denotes grouped outer-fold evaluation. Only the accepted components are deployed.}
\label{tab:falsification}
\centering
\begin{tabular}{lllll}
\toprule
Hypothesis & Experiment & Scope & Primary change & Decision \\
\midrule
More SFT fixes factor errors & Hard-error continuation & Dev & Factor MF1 $-0.1229$ & Reject \\
All factors prefer hidden states & Global layer-63 probe & Dev & MAP $+0.0251$, MF1 $+0.0002$ & Reject \\
Relative reasoning fixes boundaries & Pairwise rank / structured CoT & Dev & Factor MF1 $-0.0128$ / $-0.0056$ & Reject \\
Explicit event roles improve Task 1 & Risk event compiler & OOF & Normalized Task 1 $-0.0016$ & Reject \\
Each routed factor needs its own head & Seven independent probes & Confirm & Factor MF1 $-0.0086$ & Reject \\
Readout mismatch is label local & Seven-label latent route & Confirm & Factor MF1 $+0.0105$ & Accept \\
Final token misses risk boundary & Layer-63 risk readout & Confirm & Risk WF1 $+0.0135$ & Accept \\
\bottomrule
\end{tabular}
\end{table*}

The ablations narrow the interpretation of the gain. Extra SFT does not help, so the result is not explained by additional adaptation alone. Global replacement fails, so hidden states are not uniformly superior to token margins. Seven independent factor probes also fail, showing that the successful route depends on regularized cross-label sharing rather than unrestricted per-label capacity.

\section{Discussion}
\subsection{What Does the Latent Readout Establish?}
The result establishes predictive accessibility, not a mechanistic account of suicidal reasoning. The layer-63 state contains information from which a low-capacity classifier generalizes better than the answer-token boundary on selected outputs. Because the probes are supervised, some performance can originate in the probe itself. We limit this risk through linear models, strong regularization, grouped inner cross-validation, frozen confirmation folds, and comparison against the final-token margin. Control-task and intervention analyses would be needed to show that the same features causally drive the model's original computation~\cite{hewitt2019probes,belinkov2022probing}.

\subsection{Why Heterogeneous Readouts?}
The three output spaces impose incompatible constraints. Risk requires a global comparison among four severity cards; factors require independent decisions so that one generated label does not suppress another; evidence requires exact offsets and a variable-size set. \method{} does not force these tasks into one generation grammar. Semantic verification is shared, while the final decision rule follows the output structure. The latent components are linear models over cached states and do not require additional 27B networks.

\subsection{Psychological Interpretability}
Label cards make inclusion and exclusion boundaries explicit and inspectable. They can reveal whether a prediction follows a future appraisal (hopelessness), self-evaluation (low self-esteem), resource availability (social support), or material deprivation (low socioeconomic status). Nevertheless, the cards are operational annotations rather than validated psychometric instruments. Their predictive validity in this dataset does not establish construct validity across populations, platforms, or cultures.

\subsection{Claim Boundaries}
Table~\ref{tab:claimscope} separates what each experiment establishes from what it does not. This distinction is especially important here because a probe can expose predictive information without identifying the mechanism that produced it, and an exact copied span can be traceable without being causally sufficient for the prediction.

\begin{table*}[t]
\caption{Evidence and scope of the principal claims.}
\label{tab:claimscope}
\centering
\small
\begin{tabular}{@{}p{0.22\textwidth}p{0.32\textwidth}p{0.38\textwidth}@{}}
\toprule
Claim & Supporting evidence & Scope boundary \\
\midrule
The final-token boundary is suboptimal for risk. & Frozen Folds 1/2 both improve; pooled WF1 $+0.0135$; bootstrap probability of positive gain 0.983. & Predictive accessibility in Qwen3.8-27B on this dataset; not a causal account of internal reasoning. \\
A selective shared latent route helps factor prediction. & Aggregate MF1 improves on both confirmation folds; global replacement and independent heads fail. & Route is static and development selected; only three routed labels improve on both confirmation folds. \\
Evidence is non-fabricating and traceable. & Every phrase is returned by exact character offsets and set decoding raises row-level OOF phrase F1. & Verbatim grounding does not prove that the selected phrase is causally sufficient for risk prediction. \\
The submitted system remains competitive on the hidden test distribution. & Public Task 1/Task 2/composite scores are 0.8052/0.6636/0.7627. & One benchmark and one principal backbone; the leaderboard does not expose risk and evidence separately. \\
\bottomrule
\end{tabular}
\end{table*}

\subsection{Future Work: Learning When to Trust a Readout}
Table~\ref{tab:routediagnostic} shows that the fixed factor route is useful in aggregate but heterogeneous across labels. This suggests replacing the hand-selected route with a regularized instance--label meta-router. For post $x_i$ and label card $c_l$, let $m_{i,l}$ denote the answer-token margin and $g_{i,l}$ the shared latent-probe score. A router predicts
\begin{equation}
\pi_{i,l}=\sigma\!\left(\mathbf{u}^{\top}\boldsymbol{\phi}(x_i,c_l,m_{i,l},g_{i,l})\right),
\end{equation}
and compiles the final score as
\begin{equation}
s_{i,l}=\pi_{i,l}g_{i,l}+(1-\pi_{i,l})m_{i,l}.
\end{equation}
The meta-features $\boldsymbol{\phi}$ should be observable at test time and estimated without crossing user folds. Candidate features include post length and clause dispersion; label support, prevalence, lexical specificity, and card-embedding abstraction; token-margin magnitude and entropy; and token--latent disagreement. A hierarchical prior could share evidence across labels while shrinking rare-label routes toward the token-margin baseline. To avoid converting routing noise into overfitting, router fitting, feature normalization, and route-threshold selection would all occur inside nested user-grouped folds, with an abstention region when neither readout is reliably preferred.

The broader claim also needs external validation. We would test the same frozen protocol across multiple Qwen parameter scales and at least one bidirectional encoder, separating model-scale transfer from architecture transfer. Cross-domain evaluation on compatible CLPsych suicide-risk and evidence corpora~\cite{chim2024overview} would then test platform transfer; because their ontologies do not necessarily match the present 24 factors, risk/evidence transfer and factor-ontology transfer should be reported separately rather than forced into one score. These experiments would establish whether heterogeneous readout selection is a general property or a dataset--backbone interaction.

\section{Ethical Considerations and Limitations}
This system is a research classifier, not a clinical diagnostic or intervention tool. False negatives may overlook urgent risk, while false positives can stigmatize users or trigger inappropriate action. Reddit language is context dependent and may include quotation, sarcasm, fiction, or discussion of another person. The data distribution is not representative of the general population, and demographic performance cannot be audited because protected attributes are unavailable.

All training examples are handled under the competition Data Usage Agreement. We do not attempt to identify users, join records to external sources, or expose post text in this report. The system runs locally and uses no online LLM API at inference time. Human review, clear escalation protocols, privacy protection, and prospective clinical validation would be mandatory before any real-world use.

Several technical limitations remain. First, the sample is small for 24 factors, with only eight positives for the rarest label. Second, the selective route is label static; it cannot defer only ambiguous instances. Third, the evidence decoder guarantees exact copying but not causal faithfulness because risk prediction can read the full post. Fourth, evidence confirmation is not fully nested at the candidate-generator level. Fifth, development and confirmation use the same source dataset; the public leaderboard is an external distribution check but does not reveal component metrics. Finally, Qwen3.8-27B is computationally expensive despite 4-bit adaptation.

\section{Reproducibility}
The submitted package contains the training and inference notebooks, source modules, dependency specification, artifact map, and report source. When run, the notebooks serialize the frozen JSON configurations, fold manifests, row IDs, OOF logits, and audit tables used by later stages; these large generated artifacts and pretrained weights are not redistributed. Expensive stages are resumable at fold and scoring-chunk granularity. Each cache stores row IDs and is rejected when alignment differs. Factor retrieval is rebuilt per fold and excludes validation users and duplicate texts. We pin the meta-calibrator's scikit-learn version because serialized pipeline behavior can otherwise change across releases.

Table~\ref{tab:config} lists the main configuration. Test inference averages the available fold models and then applies only frozen probes, routes, thresholds, and evidence-set rules. No public-leaderboard labels are used for fitting.

\begin{table}[t]
\caption{Main reproducibility configuration.}
\label{tab:config}
\centering
\begin{tabular}{ll}
\toprule
Component & Setting \\
\midrule
Backbone & Qwen3.8-27B, 64 layers \\
Quantization & NF4, double quant., bf16 compute \\
Adapters & LoRA $r=16$, $\alpha=32$, dropout 0.05 \\
SFT & 1 epoch, lr $10^{-4}$, grad. accum. 32 \\
Maximum length & 1,536 tokens; left truncation \\
Risk readout & Layer 63, logistic $C=0.01$ \\
Factor readout & Layer 63, logistic $C=0.001$ \\
Factor route & 7 of 24 labels \\
Evidence meta model & Logistic $C=0.1$, threshold 0.40 \\
Outer split & 3-fold, grouped by user \\
Hardware & One NVIDIA A100 80GB \\
\bottomrule
\end{tabular}
\end{table}

\section{Conclusion}
This work formulates explainable suicide-risk detection as semantic verification followed by output-structure compilation. The same post--card computation yields both an answer-token margin and a latent answer state, but the appropriate readout depends on whether the target is ordinal, independently multi-label, or an exact span set. A layer-63 risk comparator and a seven-label shared factor route improve frozen confirmation results, whereas global latent replacement, independent factor probes, additional SFT, ranking, and structured CoT do not. Evidence is produced through exact-span selection rather than free-form generation. The combined results support a specific conclusion: some benchmark errors arise at the readout boundary even when useful predictive information remains accessible in the verifier state. The contribution is therefore not hidden-state probing alone, but a falsifiable rule for deciding when the answer token should be retained, replaced, or supplemented under different output constraints.

\section*{Acknowledgment}
We thank the IEEE BigData Cup 2026 organizers for constructing the benchmark and providing the annotated data. We also acknowledge the developers of the open pretrained models used in this work.

\balance
\bibliographystyle{IEEEtran}
\bibliography{references}

\end{document}